\documentclass[12pt]{article}
\usepackage[left=2cm, right=2cm, top=2cm, bottom=2cm]{geometry}
\RequirePackage{xspace,theorem,ifthen,calc}
\RequirePackage{amsfonts,amssymb,amsbsy}
\RequirePackage{amsmath,subcaption}
\usepackage{latexsym}
\usepackage{xcolor}
\usepackage{graphicx}
\usepackage{float}
\usepackage{caption}
\usepackage{subcaption}
\usepackage{nicefrac}
\usepackage{bm,booktabs}
\usepackage{setspace}
\usepackage[ruled,vlined]{algorithm2e}
\title{{\it Sparsistent} filtering of comovement networks from high-dimensional data\footnote{ASC acknoweldges R\&P grant from Indian Institute of Management Ahmedabad. We are grateful to Vikram Sarabhai Library for providing the data utilized in this paper. All remaining errors are ours.}}
\author{Arnab Chakrabarti\thanks{Misra Centre for Financial Markets and Economy, Indian Institute of Management Ahmedabad, Gujarat 380015, India. Email: arnab\_c@zohomail.in}
	\and
	Anindya S.~Chakrabarti\thanks{(Corresponding author) Economics Area and Misra Centre for Financial Markets and Economy, Indian Institute of Management Ahmedabad, Gujarat 380015, India. Email: anindyac@iima.ac.in}
}
\date{\today}

\begin{document}
	
	\maketitle
	
	\begin{abstract}
		
		Network filtering is an important form of dimension reduction to isolate the core constituents of large and interconnected complex systems. We introduce a new technique to filter large dimensional networks arising out of dynamical behavior of the constituent nodes, exploiting their spectral properties. As opposed to the well known network filters that rely on preserving key topological properties of the realized network, our method treats the spectrum as the fundamental object and preserves spectral properties. Applying asymptotic theory for high dimensional data for the filter, we show that it can be tuned to interpolate between zero filtering to maximal filtering that induces sparsity and consistency while having the least spectral distance from a linear shrinkage estimator. We apply our proposed filter to covariance networks constructed from financial data, to extract the key subnetwork embedded in the full sample network.
	\end{abstract}
	
	\clearpage
	\newpage
	
	\section{Introduction}

	Network representation of large dimensional complex systems has become a standard methodology to delineate
	the nature of linkages across a large number of constituent entities comprising the systems \cite{newman2003structure}.
	Examples range across systems varying widely in terms of nature and architecture: economic and financial networks \cite{bougheas2015complex,battiston2016complexity}, social networks \cite{vega2007complex}, biological networks like food webs \cite{williams2000simple}, technological networks like world wide web \cite{huberman1999growth} and transportation networks \cite{sen2003small} among many others.
	Broadly speaking, there are two major strands of literature that starts from the analysis of the realized network. One strand of the literature utilizes networks to explore dynamics on it \cite{newman2006structure}, using the realized network as the true representation of the linkages.
	The other literature goes backward to extract true linkages from the realized linkages \cite{barfuss2016parsimonious,radicchi2011information}, maintaining the idea that some of the realized linkages in fact might be spurious. We are interested in the second stream of literature where the fundamental objective is to isolate and filter the key subnetwork out of a large dimensional realized network.

	In a complex dynamical system, the correlation matrix of time-varying responses of the constituent entities captures pairwise-linkages between the entities. A co-movement network is constructed by considering each response variable as a node of the graph and an undirected edge between two nodes exists if the corresponding correlation is nonzero. 
	This kind of network construction out of observational multi-variate data has been very successful as a modeling paradigm in finance \cite{marti2017review} and biology \cite{cho2012network,barabasi2004network} among others. However, such inference about existence of linkages from purely observational data has a problem.
	As the pairwise sample correlation is hardly equal to 0 (even when the true correlation is 0), the realized co-movement network will always be a complete graph. The size of the correlation matrix grows as square of the number of nodes. Therefore for real-life data, a complete graph constructed from a such a large correlation matrix might have edges carrying information that would be spurious in nature. Many of the edges, particularly the edges with very low correlation, contain very little information and a likely scenario is that they lead to {\it false discovery} of linkages.
	Hence, before carrying out the statistical analysis of a co-movement network, it is important to extract only the meaningful interactions or correlations. Prominent network filtering techniques, like \emph{minimum spanning tree} \cite{mantegna1999hierarchical} (MST) or \emph{planar maximally filtered Graph} (PMFG) \cite{tumminello2005tool}, aspire to do so by reducing the graph to a subgraph containing the maximum amount of information regarding the system's collective behavior by preserving geometric properties of the realized network (connectivity in case of MST and closed loops with three or four nodes in case of PMFG). The second type of filtering emphasizes the statistical significance of edges \cite{marcaccioli2019polya}. The third type of filtering focuses on the spectral structure \cite{hermsdorff2019unifying}.

	In this paper, we propose a new filtering technique for large networks constructed from high-dimensional data, utilizing the spectral properties. 
	Drawing from statistical theory of high-dimensional covariance matrix estimation, we develop a flexible method to find out the sparse adjacency matrix that represents the key subnetwork of the full network. Theoretically, the filtered network retains maximum similarity with the true spectral structure. The method is quite flexible as it allows the tuning the degree of filtering within a range of zero to maximal permissible pruning of edges.  
	Two important properties of the filter are as follows: first, the filter generates {\it sparsity} in the covariance matrix which makes the filtering possible, and secondly, the filter statistically consistently prunes spurious linkages leading to reduction on false discovery of linkages. The combination of these two properties lead to the {\it sparsistence} of the resultant filter.

	Fundamentally, our approach depends on the literature on large dimensional covariance matrix estimation. For a large interacting system, a comovement network ia a high-dimensional graph. Often the number of nodes is in the order of
	number of observations leading to a well recognized problem that the eigenvalues of the sample covariance matrix do not converge to their population counterpart~\cite{marvcenko1967distribution}. This result dictates the fact that the sample covariance matrix is not an consistent estimate of the true covariance matrix~\cite{pourahmadi2013high, davis1970rotation}. Therefore, efficient estimation of high-dimensional covariance matrix is a relevant problem in context to large network analysis. A broad class of well-conditioned shrinkage/ridge-type
	estimators were proposed to circumvent the problem~\cite{ledoit2004well}. Element-wise regularization methods were also proposed to achieve sparsity \cite{bickel2008covariance, bickel2008regularized, rothman2008sparse, bickel2004some}. Some of the methods require a natural ordering among the variables \cite{rothman2010new, bickel2008regularized}. Some of the proposed estimators fail to guaranty positive definiteness. Some form of \emph{tapering} matrix~\cite{furrer2007estimation} and maximum likelihood estimator under positive definite and sparsity constraints~\cite{chaudhuri2007estimation} were proposed to ensure positive definite covariance estimator.  
	We borrow the idea of consistent estimation of sparse covariance matrices from this literature.
	
	However, there are two ways of approaching the problem of inference of linkages. In this paper we deal with the graph implied by the covariance matrix and utilize a non-parametric approach.
	In particular, we leverage the properties of regularized covariance estimators in \cite{ledoit2004well} and \cite{rothman2009generalized}.
	The complementary approach is through the application of graphical Lasso algorithm for Gaussian graphical model \cite{cai2011constrained, friedman2008sparse}. However, graphical Lasso algorithm attempts to estimate the precision matrix and not the covariance matrix. Despite it's popularity, graphical Lasso algorithm is a strongly parametric approach and applicable within a rather restricted class of models. 
	
	Several attempts have been made to develop filtering methods while preserving large scale structure \cite{hamann2016structure, hermsdorff2019unifying}. \cite{hamann2016structure} show that local filtering techniques can preserve network properties more that global filtering methods and propose a new sparsification technique that preserves edges leading to nodes of local hubs. \cite{imre2020spectrum} integrate spectral clustering and edge bundling for effective visual understanding.  Under some distributional assumptions, statistical methods have been proposed to extract the backbone of the network \cite{coscia2017network, marcaccioli2019polya}. These works develop statistical tests for significance of edges. Some methods are proposed to find the irreducible backbone of a network from a sequence of temporal contacts between vertices \cite{kobayashi2019structured}.  
	
	Finally, we note that the proposed filter is more efficient than the filters based on random matrix theory, as those filters lead to shrinkage of all elements in the correlation matrices due to spectral decomposition without converting any of them to zero. Thus the resultant network is of the same dimension as the original network \cite{sinha2010econophysics}. There are application of hard thresholding on the resultant network to reduce the size of the network by removing edges with low weights. However, such a technique is fundamentally ad-hoc as there is no intrinsic property that can fix the threshold \cite{sinha2010econophysics}.
	In the present context, we avoid both problems by essentially targeting consistent covariance matrix estimator via sparsity and the distance between the eigenspectra of the target and the filtered matrix uniquely pins down the degree of thresholding and consequently, the degree of filtering.

	The rest of the paper is organized as follows: Section \ref{section:background} introduces the essential notations used throughout and discusses necessary statistical background. Readers familiar with high-dimensional covariance matrix estimation problem can skip this part and can directly go to the next section~\ref{sec:proposed_algorithm}. In section \ref{sec:choices_n_alternatives}, we present some of the possible alternatives of the choices we made in the algorithm. We have presented applications of the filter to real-life data in section \ref{sec:real_data}. Section \ref{sec:summary} summarizes the paper and concludes.

	\section{Notations and Technical Background on Covariance Matrix Estimation}
	\label{section:background}

	Throughout this paper we maintain the following notations: 
	\begin{itemize}
		\item $\mathcal{D}$: the $n\times p$ data matrix consisting of $p$ variables and $n$ independent observations. 
		\item $\Sigma$: true (unobserved) covariance matrix ($p\times p$) of $p$ variables. 
		\item $S$: sample covariance matrix of size $p\times p$ calculated from data matrix $\mathcal{D}$. 
		\item $S_{LW}$: Ledoit-Wolf estimator of size $p\times p$ of covariance matrix $\Sigma$. 
		\item $S_{\eta}$: Thresholded (sample) covariance matrix of size $p\times p$ corresponding to the threshold $\eta$. 
		\item $S_{\eta^*}$: Maximally filtered covariance matrix of size $p\times p$ for optimally chosen threshold parameter $\eta^*$ with zero cost for filtering. 
		\item $S_{\tilde{\eta}}$: Tuned filtering of covariance matrix of size $p\times p$ for optimally chosen threshold parameter $\tilde{\eta}$ for positive cost for filtering. 
		\item $\Gamma(S_{\eta})$: Network corresponding to $S_{\eta}$.
	\end{itemize}   
	Following the above notation, the goal of our proposed methodology is to find an optimal threshold parameter $\eta^*$
	such that the corresponding filtered network $\Gamma(S_{\eta^*})$ will have the {\it sparsistence} property. Below we define all concepts and discuss each of the steps in detail.

\begin{figure}
		\begin{center}
			\begin{minipage}{0.3\textwidth}
				\includegraphics[width=\textwidth]{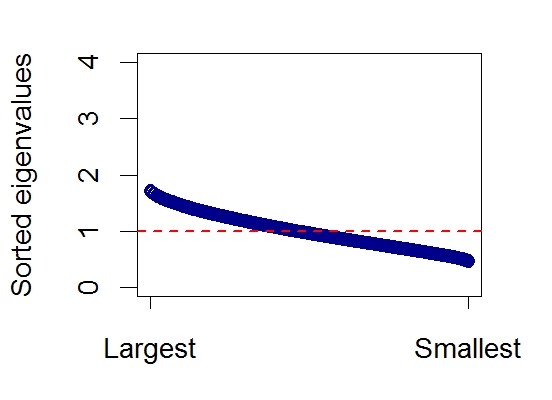}
				\vspace{-.7cm}
				\subcaption{p/n=0.1}
			\end{minipage}
			\begin{minipage}{0.3\textwidth}
				\includegraphics[width=\textwidth]{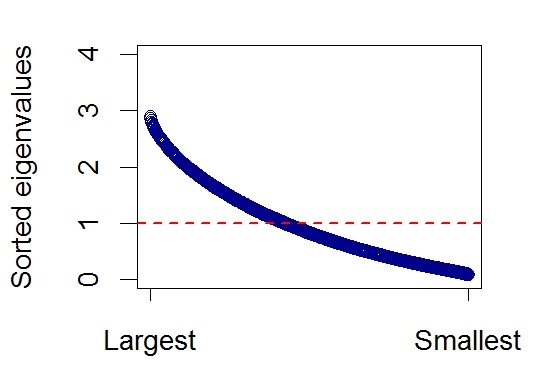}
				\vspace{-.7cm}
				\subcaption{p/n=0.5}
			\end{minipage}\\
			\begin{minipage}{0.3\textwidth}
				\includegraphics[width=\textwidth]{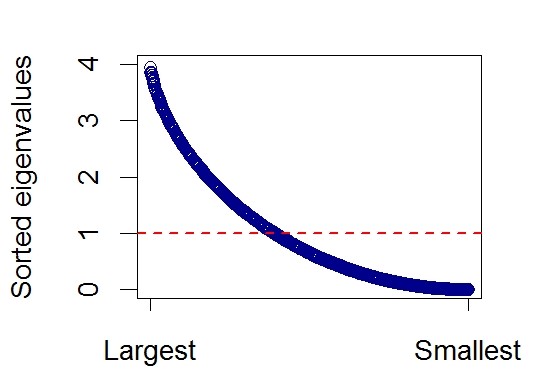}
				\vspace{-.7cm}
				\subcaption{p/n=1}
			\end{minipage}
			\begin{minipage}{0.3\textwidth}
				\includegraphics[width=\textwidth]{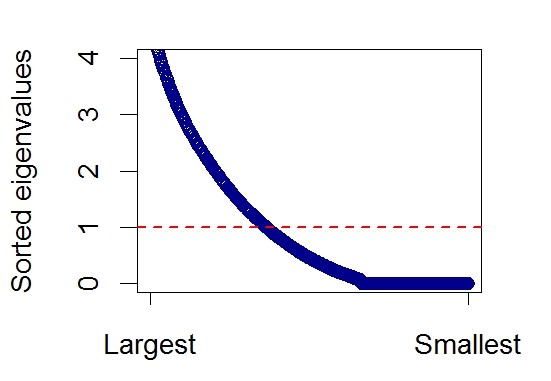}
				\vspace{-0.7cm}
				\subcaption{p/n=1.5}
			\end{minipage}
			\caption{Plot of sorted eigenvalues corresponding to the true (dotted line) and sample covariance matrix (solid line) with $n$ draws from a $p$ dimensional normal distribution with covariance matrix $I$ (identity matrix). Since all eigenvalues of an identity matrix would be equal to one, the true spectrum is shown as a horizontal line at 1. Each panel represents a specific dimension to sample size ratio varying from 0.1 to 1.5. Higher $p/n$ ratio leads to larger deviation of the sample spectrum from the true spectrum \cite{pourahmadi2013high}, which necessitates spectral shrinkage as described in Sec. \ref{sec:Stein} and \ref{sec:Ledoit_wolf}.
			}\label{eigenvalue_plot}
		\end{center}
	\end{figure}

	With multivariate data, the population covariance matrix is estimated by its sample counterpart. The sample covariance matrix has unbiasedness and other useful large sample properties \cite{bai2010spectral}. However, these properties are established under the assumption that the number of observations is large while the number of variables being constant. The difference between the multivariate statistical theory and high-dimensional statistics is that the latter considers the case where the number of variables ($p$) also grows with the number of observations ($n$). Under such assumption, the sample covariance matrix does not behave desirably and becomes inconsistent. When sample is drawn from a high-dimensional Gaussian distribution with true covariance matrix $I$, the difference between the true and sample spectra increases with the dimension to size ratio- as illustrated in Fig.~\ref{eigenvalue_plot}. This fact is theoretically proved by \emph{Mar{\v{c}}enko-Pastur} theorem and consequent developments \cite{bai2010spectral}. For this reason, several attempts have been made to construct more efficient estimator of high-dimensional covariance matrix. Here, we will describe a few of these approaches which are relevant to this paper and used in section~\ref{sec:proposed_algorithm} to develop the algorithm. 

	\subsection{Stein's approach}
	\label{sec:Stein}
	
	Fig.~\ref{eigenvalue_plot} shows that under high-dimensional setup, the eigenvalues of the sample covariance matrix deviates considerably from their population counterparts. However, the problem itself suggests a possible way out. We can see (Fig.~\ref{eigenvalue_plot}) that as the dimension to sample size ratio goes up, the sample spectra move further away from the true spectra. So shrinking the eigenvalues towards a central value may lead to a better estimator.  
	Such a strategy was suggested by Stein \cite{stein1956inadmissibility} and the proposed covariance estimator takes the following form: 
	\begin{equation}
	\hat{\Sigma}=\hat{\Sigma}(S)=P\psi(\Lambda)P', \label{eq:stein}
	\end{equation}
	where the spectral decomposition of $S$ is given by $S=P\Lambda P'$, with $\Lambda=\mathrm{diag}(\lambda_{1},\lambda_{2},...,\lambda_{p})$ being the diagonal matrix of eigenvalues of $S$ and $P$ being the matrix of eigenvectors; $\psi(\Lambda)= \mathrm{diag}(\psi(\lambda_{1}),\psi(\lambda_{2}),...,\psi(\lambda_{p}))$ is also a diagonal matrix. If $\psi(\lambda_{i})=\lambda_{i}\forall i$
	then $\hat{\Sigma}$ is the usual sample covariance matrix $S$. $\psi$ shrinks the eigenvalues $\lambda$ and thus reduce the deviation from its true counterpart. Clearly, this approach only regularizes the eigenvalues and keep the eigen vectors of the sample covariance matrix unaltered. Due to this reason this type of estimators are also called \emph{rotation equivariant} covariance estimator.

	\subsection{Ledoit-Wolf estimator}
	\label{sec:Ledoit_wolf}
	
	A problem with Stein's original prescription is that it does not ensure monotonicity and nonnegativeness of the eigenvalues \cite{pourahmadi2013high}. This problem had been addressed by Ref. \cite{ledoit2004well} which formulated a general approach towards shrinkage by defining a rotation equivariant regularization based on the following minimization problem: 
	\begin{equation}
	\underset{\Psi}{\text{min}}\|P\Psi P'-\Sigma\|
	\label{eq:regularization_optim}
	\end{equation} 
	where $\|.\|$ can be any matrix norm. Most widely considered norm is the Frobenius norm.\footnote{The Frobenius norm of an arbitrary matrix $A$ of order $r\times m$ is $\|A\|^F=\sqrt{\text{tr}(AA')/r}$. \label{foot:Frobenius}} 
	
	A particularly useful solution for the optimal $\psi$ was proposed by Ledoit and Wolf \cite{ledoit2004well} which is based on the observation that
	the sample covariance matrix is an unbiased estimator of the population covariance matrix. This fact remains true for high-dimensional data as well. But in high-dimensional setup, the sample covariance matrix becomes considerably unstable i.e. the deviation from the true covariance matrix can potentially be large. On the other hand if we use a structured covariance estimator- such as an identity matrix then the estimator, while being severely biased under misspecification of the structure, will have very little variability. They showed that a suitably chosen linear combination of these two types of estimators would outperform each of them where the coefficients/weights of linear combination is chosen to optimize the bias-variance trade off. Formally, the Ledoit-Wolf estimator $S_{LW}$ \cite{ledoit2004well} is defined as
	\begin{equation}
	S_{LW}=\alpha_{1}I+\alpha_{2}S
	\label{eq:Ledoit_Wolf}
	\end{equation} 
	where $I$ is a $p\times p$ identity matrix and $\alpha_1$ and $\alpha_2$ are chosen to minimize the risk corresponding to the loss function $p^{-1}\mathrm{tr}(S_{LW}-\Sigma)^2$.
	\subsubsection{Consistency of Ledoit-Wolf estimator}
	\label{sec:consistency_LW} 
	
	Since $S$ is positive definite, $S_{LW}$ can also be shown to be positive-definite and consistency of such estimator depends on the growth rate of $p$, the moments and the association structure of the data \cite{ledoit2004well}. 
	More precisely, the $p(p+1)/2$ elements of the true covariance matrix can be consistently estimated if three conditions hold described below. 
	
	In large dimensional covariance matrices, both $p$ and $n$ grows. Therefore it is a common practice to write $p$ as $p_n$ (function of $n$) and to consider $n$ in the limit. Let us define $Y=\mathcal{D}V$, where $\mathcal{D}$ is a $n\times p$ matrix and $V$ is the matrix whose columns are the normalized eigenvectors of $\mathcal{D}$.  
	Denote the $i$th entry of any row by $Y_i$. Also, let us denote the set of all quadruples made of four distinct elements of $\{1,~2,~3,~..,~p\}$ as $Q_n$. 
	
	The three conditions are the following: 
	\begin{enumerate}
		\item[C1:] There exists a constant $K_1$ independent of $n$ such that $p_n/n\leq K_1$. 
		\item[C2:] There exists a constant $K_2$ independent of $n$ such that $\sum_{i=1}^{p_n} E(Y_i)^8<K_2$.
		\item[C3:] \[\underset{n\rightarrow \infty}{\mathrm{lim}}\frac{p_n^2}{n^2}\times\frac{\sum_{(i,j,k,l)\in Q_n}(Cov(Y_iY_j,Y_kY_l))^2}{\#Q_n}=0,\] where $\#Q$ denotes the cardinality of the set $Q$.   
	\end{enumerate}
	C1 says that $p$ can either remain constant or grow with $n$. That means this method cannot be used (more specifically consistency cannot be achieved) for data for which $p>>n$.         
	
	\subsection{Sparsity and threshold estimator}
	\label{sec:sparsity_threshold_estimator}
	Threshold estimator of high-dimensional covariance matrix regularizes both eigenvalues and eigenvectors as opposed to the Ledoit-Wolf estimator which only regularizes the eigenvalues of the sample covariance matrix. Threshold estimator is particularly useful when the true covariance matrix from the data generating process is \emph{sparse}, i.e. many of the non-diagonal entries of the covariance matrix are 0 or close to 0. This assumption is reasonable for a wide range of practical scenarios. Threshold estimator forces all the off-diagonal entries below a suitably chosen threshold to 0. Even if the corresponding entries of the true covariance matrix are nonzero, the threshold estimates of those entries entail only a bias but no dispersion as estimated by a fixed constant which is zero. 
	
	The objective function would be the Frobenius norm (see footnote \ref{foot:Frobenius}) of the difference between the thresholded matrix and empirical covariance matrix obtained from repeated sampling \cite{bickel2008covariance}. If the threshold is large, this method produces a sparse covariance matrix. We will denote the threshold matrix by $S_{\eta}$, where $\eta>0$ is the chosen threshold and $S$ is the usual sample covariance matrix; 
	\begin{eqnarray}
	S_{\eta}&\equiv& [s_{\eta}({i,j})] \nonumber\\ 
	&\equiv& [s_{\eta}(i,j)I(|s(i,j)|<\eta)]\quad ~~~~~~~~~~~
	\label{eq:threshold_est}
	\end{eqnarray}
	where $I(.)$ is the indicator function. So the entries of $S$ which are less than $\eta$ in magnitude are replaced by 0. 
	The optimal threshold parameter $\eta$ can be chosen by cross validation. 
	
	The resulting threshold estimator would be consistent under the assumption $\log(p)/n\rightarrow 0$, and it is shown to be uniform for a class of matrices satisfying a condition that captures a notion of ``approximate sparsity". One problem of such estimator is that it does not always preserves positive definiteness \cite{bickel2008covariance}.
	
	Threshold estimators can be further extended to a broader class of matrices called \emph{generalized thresholding operators}, which combines two regularization methods: thresholding and shrinkage \cite{rothman2009generalized}. When the true covariance is sparse, generalized thresholding estimators can identify the the true zero entries with probability 1. This property commonly called as \emph{sparsistency}. The sufficient conditions required to achieve this are the following \cite{pourahmadi2013high}: 
	\begin{enumerate}
		\item[C1:] The data generating process is Gaussian. 
		\item[C2:] The variances are bounded above by a constant, i.e. $\sigma_{i,j}\leq C$ for a sufficiently large $C$. 
		\item[C3:] $C\sqrt{\frac{\text{log}~ p}{n}}=o(1)$.
	\end{enumerate}
	
	\section{Sparsistent Filtering of Networks}
	\label{sec:proposed_algorithm}

	Fundamentally, our objective is to combine the feature of sparsity from the large dimensional covariance matrix estimation along with preservation of the underlying network topology. We elaborate on these two related but separate features below. 
	
	Intuitively, the problem of filtering a complex network is equivalent to the problem of deleting a number of ``less-important'' edges from the original graph such that it become less complicated and reveals an underlying structure. Therefore sparsity is an essential property of the adjacency matrix of a filtered network. Evidently, this can be achieved by a threshold estimator with appropriately chosen threshold parameter. However, in the context of network filtering, the choice of a threshold is often ad-hoc and suffers from lack of robustness \cite{sinha2010econophysics}.
	Therefore, the technical problem is what can be an efficient method for choosing the threshold that retains sparsity but is also statistically robust? One candidate would be cross-validation \cite{qiu2019threshold}. However, a threshold implied by cross-validation is purely numerical in nature and fully dependent on the realized covariance matrix where the realized covariance matrix is itself a random sample.
	As a direct implication, such a threshold does not allow inference on the underlying structure of the true covarince matrix (and therefore, the resulting network structure). Within a certain restricted class of data generating process, a threshold estimator can indeed be consistent \cite{bickel2008covariance}. However, the corresponding restrictions are too severe for direct applications to real-life data (e.g. the assumption of multivariate Gaussianity is often rejected in systems exhibiting large fluctuations, like in the case of stock market data \cite{sinha2010econophysics}). Additionally, a cross-validated threshold estimator would exhibit theoretical consistency only when the true covariance matrix is sparse to begin with. However, a network filter should be flexible enough to consider a scenario where the linkages are of small strength, but non-zero.
	
	A further question arises here about the best way to capture the network topology. One can consider observable geometric properties as well as the spectral structure of a network. There are filters which focus on the geometric properties (like connectivity in case of minimum spanning tree or closed loops in case of planar maximally filtered graphs \cite{tumminello2005tool}). However, in the present context, the spectral structure of networks is the best candidate: one, the spectral structure by definition captures network topology, and two, it is amenable to asymptotic theories and links naturally with covariance matrix estimation.

	In sum, our goal is to find a way to retain the feature of sparsity in one hand (to make the estimator efficient) while preserving the network topology on the other. We require the filter should be flexible enough to interpolate between the two.
	We achieve it by combining the Ledoit-Wolf estimator as the target for retaining statistical consistency and imposing a threshold estimator that emulates the corresponding spectral structure. We show that the resulting filter inherits both sparsity as well as consistency.

\begin{figure}
		\begin{center}
			\includegraphics[scale=0.75]{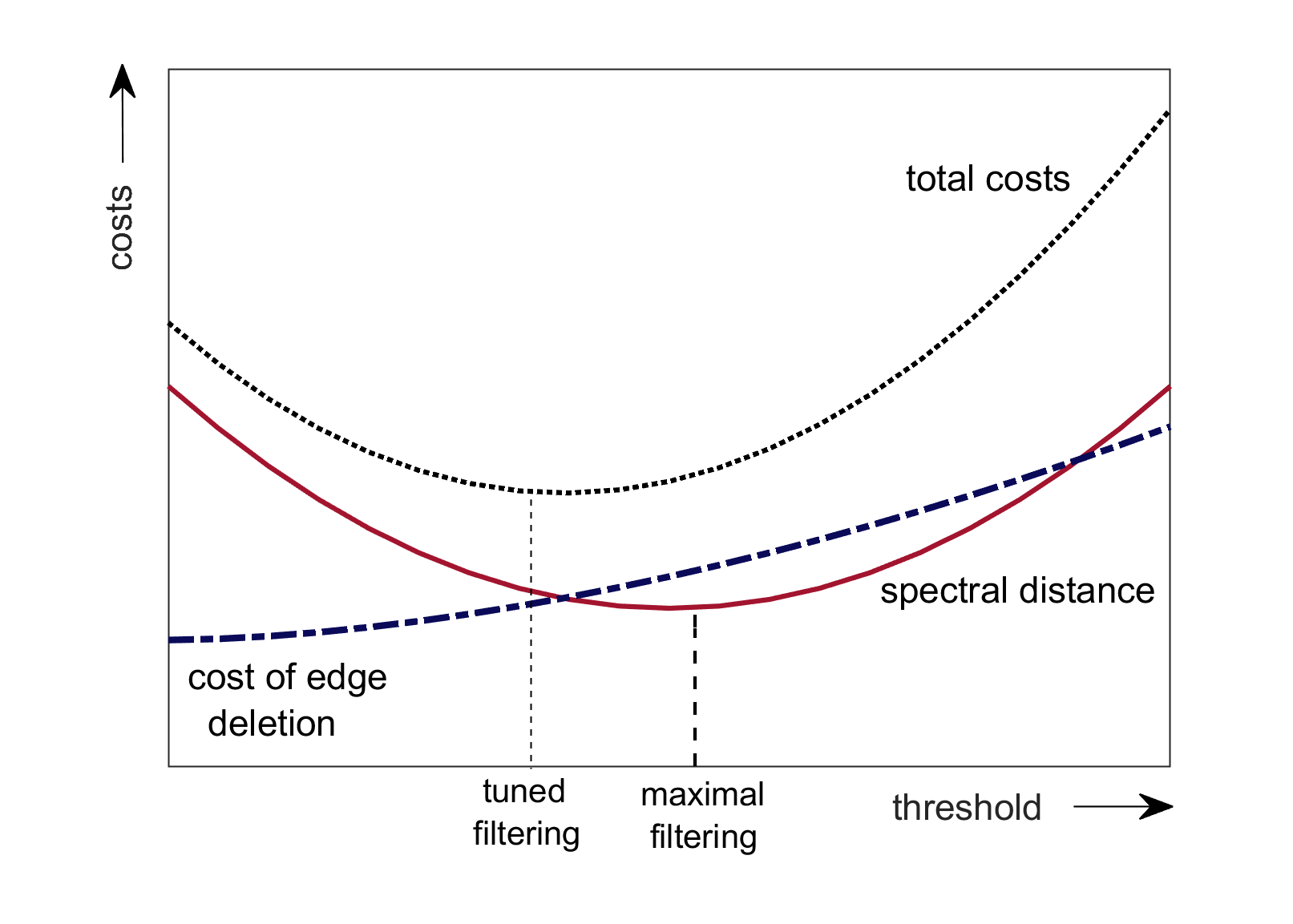}
			\caption{Illustration of the distance function and cost function against the values of $y(\eta)$- the number of deleted edges. The x-axis represents values of $y$, the black and red curves stand for $d(F^{S_{\eta}},F^{S_{LW}})$ (Eq.~\ref{eq:dist_defn}) and $C(\eta)$ (Eq.~\ref{eq:cost function}).}
			\label{fig:cost_illustration}
		\end{center}
	\end{figure}

	We explain the main idea in Fig. \ref{fig:cost_illustration}. The red thick line represents
	spectral distance of the candidate thresholded matrix from a target matrix by increasing the threshold from zero to a large enough value (we will provide the analytical details below). This distance represents a cost, indicating that a higher distance is less efficient. The distance as a function of the threshold is non-monotonic in nature. By increasing the threshold, initially the distance between the thresholded matrix from the target spectrum reduces and beyond a level, further increase in the threshold leads to an increase in the distance. The global minimum here corresponds to what we call {\it maximal filtering}, which is obtained by simply minimizing the spectral distance. 
	
	However, we have to also consider a case where the true network is not necessarily sparse and there can be edges which can be of small magnitude. 
	Thus in a general context, deleting them would entail a cost. 
	We note that for a small threshold the edges being deleted would have small weights. But as the threshold is increased, we will filter edges with larger weights. This idea can be captured through a convex cost function, which indicates that as the threshold increases, the cost associated with deletion of edges with higher weights also increases.
	The blue dashed line shows a stylized cost curve. Therefore, the proper objective function is to minimize the total cost (adding spectral distance and the cost of edge deletion) indicated by the black dotted line. The final tuned filter would extract a threshold which is less than the threshold for maximal filtering as shown on the $x$-axis.

	The algorithm described below explains the steps.  
	\subsection{Sequential steps of the algorithm}
	
	Given the data matrix $\mathcal{D}$ of size $n\times p$, our algorithm goes through the following steps and return a thresholded matrix $S_{\tilde{\eta}}$ of size $p\times p$.
	
	\begin{enumerate}
		\item {\bf Sample Covariance Matrix Construction:} From the data $\mathcal{D}$, we calculate the sample covariance matrix $S$ of size $p\times p$. 
		
		\item {\bf Construction of Ledoit-Wolf estimator:} From the sample covariance matrix $S$, we calculate Ledoit-Wolf estimator $S_{LW}$ (following Eqn.~\ref{eq:Ledoit_Wolf}).\footnote{For numerical implementation in R, we used the R package RiskPortfolios for Ledoit-Wolf matrix calculation (https://github.com/ArdiaD/RiskPortfolios) with type `oneparm'. For numerical implementation in Matlab, we have used the code titled `cov1para.m' obtained from the code repository of Ledoit-Wolf estimator (https://www.econ.uzh.ch/en/people/faculty/wolf/publications.html\#9).}
		
		\item {\bf Finding the Spectrum of Ledoit-Wolf estimator:} Eigenvalue decomposition of the covariation matrix\footnote{We choose to decompose the covariance matrix to obtain the spectrum. However one can certainly perform the identical analysis on the correlation matrix as well. Sometimes covariance can be too small and can exhibit some computational problem while implementation. Therefore we suggest it to be used on the correlation matrix.} $S_{LW}$ gives us the spectrum denoted by $\lambda(S_{LW})$, which is a $p$ dimensional vector comprising $p$ eigenvalues. The empirical distribution function of $\lambda(S_{LW})$ is denoted by $F^{S_{LW}}$.    
		
		\item {\bf Quantifying the Spectral Distance:} The goal is to find a sparse thresholded matrix that is proximate to $S_{LW}$ in terms of spectrum. We define the spectral distance  
		between two matrices $A$ and $B$ as $\text{d}(F^A,F^B)$ as the Euclidean distance between two spectra $\lambda(A)$ and $\lambda(B)$:   
		\begin{equation}
		\text{d}(F^A,F^B)\equiv d(\lambda(A),\lambda(B))=\big(\sum_{i=1}^p\big|\lambda_{i}^A-\lambda_{i}^B\big|^2\big)^{1/2}.  \label{eq:dist_defn}  
		\end{equation}
		
		\item {\bf Maximally Filtered Network:} In this step, we obtain the ``strong'' or the ``maximal'' filter $S_\eta^*$ corresponding to a threshold $\eta^*$ that has the least spectral distance from $S_{LW}$. Formally, $S_{\eta^*}$ can be obtained by minimizing the distance (Eqn.~\ref{eq:dist_defn}) of the resultant thresholded matrix from the Ledoit-Wolf matrix:   
		\begin{equation}
		\eta^*=\underset{\eta}{\text{argmin}}\{d(F^{S_{\eta}},F^{S_{LW}})\}.\label{eq:strong_filter_theshold}
		\end{equation}
		Suppose, the number of edges being deleted in $S$ to reduce the matrix to $S_{\eta^*}$ is $y^*$.\footnote{$y^*$ is a function of the threshold $\eta^*$. A higher threshold $\eta^*$ leads to deletion of a higher number of edges, implying that $y^*$ would also be higher.} Formally, we write   
		\begin{equation}
		y^*=\underset{i\neq j}{\sum}I(S_{\eta^*}(i,j)= 0) \label{eq:maximal_y}
		\end{equation}
		where $I(.)$ is the identity function and $S_{\eta^*}(i,j)$ is the $(i,j)$th entry of $S_{\eta^*}$.
		
		\item {\bf Tuned Filtering with Costly Edge deletion:} Now we impose a cost for deleting edges.
		The cost-adjusted optimal edge filtering leads to the following optimization: 
		\begin{equation}
		\tilde{\eta}=\underset{\eta}{\text{argmin}}\{d(F^{S_{\eta}},F^{S_{LW}}) + C(\eta)\}.\label{eq:optimal_filter_y} 
		\end{equation}
		where $C(\eta)$ denotes the cost of deleting $y$ edges by implementing threshold $\eta$. For practical implementation, it is easier to work with the cost function on the edges to be deleted ($C(y)$) rather than the threshold ($C(\eta)$). 
		
		A natural requirement for $C$ as a function of the number of edges to be filtered, is that it should be non-negative, continuous and potentially increasing in the first derivative leading to convexity. 
		A typical candidate for a flexible functional form of $C$ is as follows\footnote{The parameters $\theta_1$ and $\theta_2$ in the cost function given by Eqn. \ref{eq:cost function} has to be specified by the user depending on the problem and the context of application.}:
		\begin{equation}
		C(y) = \theta_1 y^{\theta_2} \hspace{.3cm} \text{where} \hspace{.3cm} \theta_{1}\ge 0,~ \theta_{2}> 1~ \text{and}~ y\equiv y(\eta).  
		\label{eq:cost function}
		\end{equation}
		Eqn.~\ref{eq:optimal_filter_y} can be equivalently written in terms of deleted edges:  
		\begin{equation}
		\tilde{y}=\underset{y}{\text{argmin}}\{d(F^{S_{\eta}},F^{S_{LW}}) + C(y)\}.\label{eq:optimal_filter_y_edge} 
		\end{equation}
		The resulting {\it tuned} threshold is $\tilde{\eta}$, number of edges deleted is $\tilde{y}$ and finally, the thresholded matrix is $S_{\tilde{\eta}}$.
		
		\item {\bf Tuned Filtered Network with Sparsistence:} We create the filtered network $\Gamma_{\tilde{\eta}}$ adjacency matrix from $S_{\tilde{\eta}}$. An edge is present if the corresponding element in $S_{\tilde{\eta}}$ is nonzero.\footnote{The covariances would not represent a metric since they can be negative. If we want to visualize the network in the metric space, we can convert covariances into correlations by dividing each covariance entry by the product of sample standard deviations of the corresponding pair of nodes, and these correlations can be transformed into a metric by using the transformation $\gamma_{ij}=\sqrt{2(1-\rho_{ij})}$ where $\rho_{ij}$ is the correlation between $i$ and $j$-th nodes \cite{sinha2010econophysics} obtained from $S_{\eta}$ for any given $\eta$.}  
	\end{enumerate}   
	
	By construction, the maximally filtered network $\Gamma(S_{\eta^*})$ would be a subset of 
	the tuned filtered network $\Gamma(S_{\tilde{\eta}})$. $\Gamma(S_{\tilde{\eta}})$ in turn would be a subset of the original network $\Gamma(S)$ corresponding to the sample covariance matrix $S$. \textcolor{black}{In our empirical studies, we have seen that the threshold $\eta^*$ is typically higher than the threshold chosen by the conventional threshold-estimator which is based on cross-validation (see Sec. \ref{sec:sparsity_threshold_estimator}). This implies that the sparsistency property (see Sec.~\ref{sec:sparsity_threshold_estimator} for sufficient conditions) will be maintained by maximal filtering because edges deleted for a threshold will also be deleted for all higher thresholds. In other words all spurious edges will be removed with probability one.}

	\subsection{An illustrative example}
	\label{sec:illustrative_example}

	\begin{figure}
		\begin{center}
			\begin{minipage}{0.4\textwidth}
				\includegraphics[trim=15 75 16 16,clip,width=\textwidth]{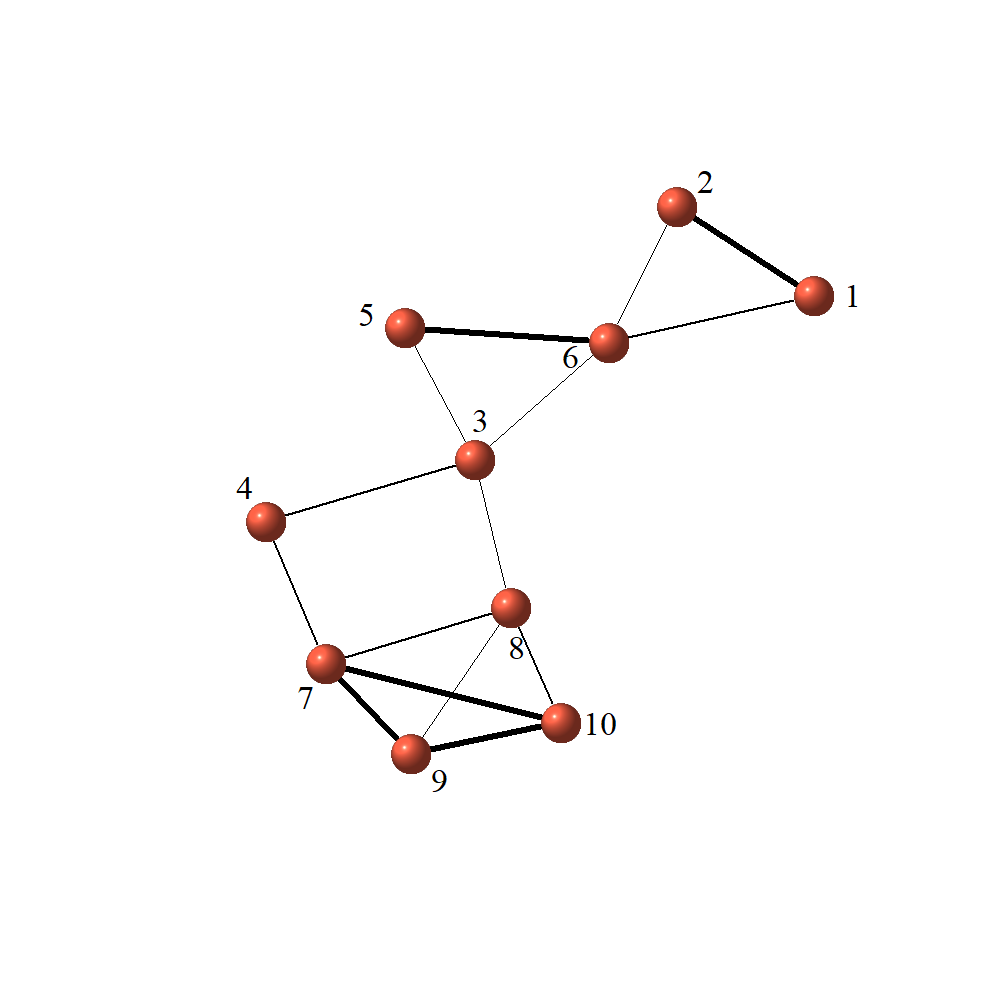}
				\vspace{-.2cm}
				\subcaption{True network: $\Gamma(\Sigma)$}\label{Fig:true_graph}
			\end{minipage}
			\begin{minipage}{0.4\textwidth}
				\includegraphics[trim=15 70 16 16,clip,width=\textwidth]{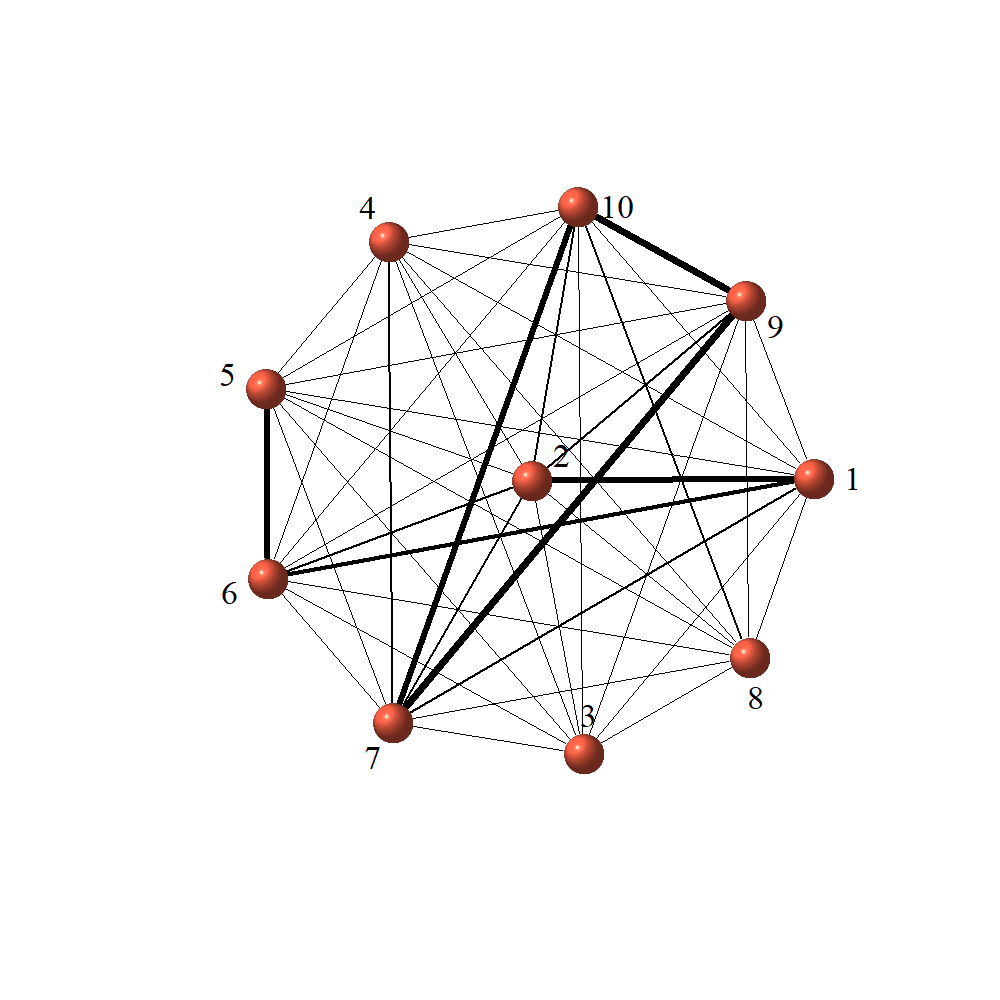}
				\vspace{-.1cm}
				\subcaption{Realized network: $\Gamma(S)$}\label{Fig:sample_graph}
			\end{minipage}\\
			\begin{minipage}{0.4\textwidth}
				\includegraphics[trim=15 60 16 16,clip,width=\textwidth]{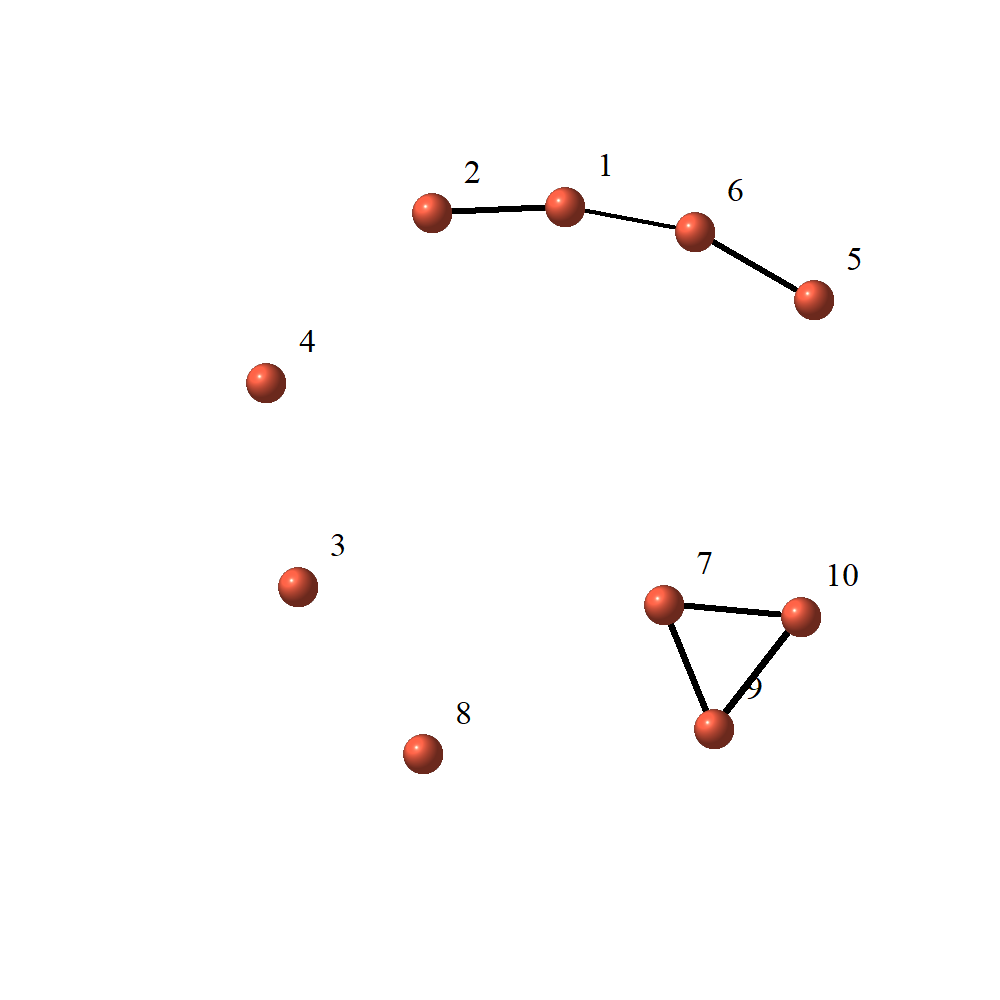}
				\vspace{-0.2cm}
				\subcaption{Maximally filtered network: $\Gamma(S_{\eta^*})$}\label{Fig:no_cost_network}
				\vspace{.5 cm}
			\end{minipage}
			\begin{minipage}{0.4\textwidth}
				\includegraphics[trim=15 80 16 16,clip,width=\textwidth]{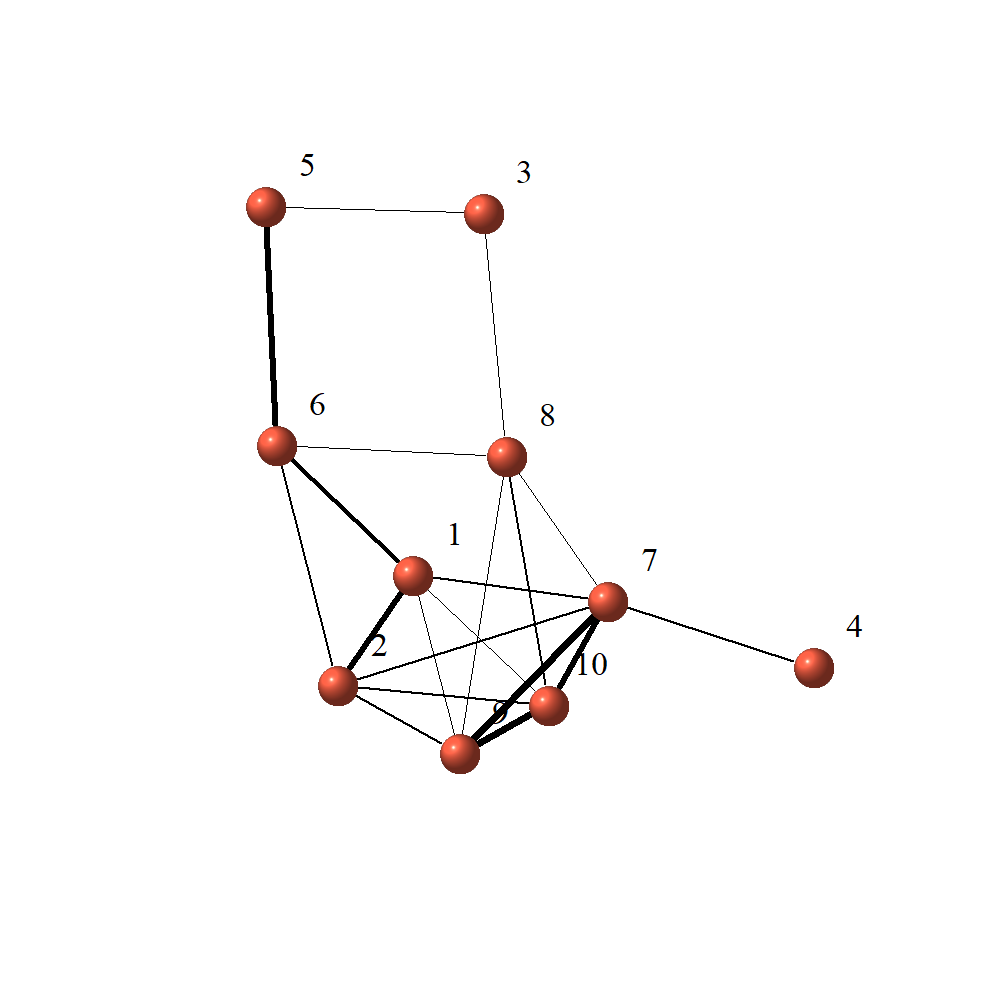}
				\vspace{-.2cm}
				\subcaption{Network with tuned filtering: $\Gamma(S_{\tilde{\eta}})$}\label{Fig:high_cost_network}
			\end{minipage}
			\caption{Illustration of the proposed algorithm. Panel (a): True sparse network $\Gamma(\Sigma)$ for simulation study with 15 edges and 10 nodes. Panel (b): The realized network $\Gamma(S)$ from simulated data corresponding to the sample covariance matrix $S$. Due to finite sampling ($n=50$), the network is fully connected with 45 edges (for $p$ = 10, the maximum number of edges $p(p-1)/2$ = 45). Panel (c): The maximally filtered network $\Gamma(S_{\eta^*})$ when $\eta^* = 0.499$, with 6 edges. Panel (d): The tuned filtered network $\Gamma(S_{\tilde{\eta}})$ where the optimally chosen threshold $\tilde{\eta}$ is 0.170. $\Gamma(S_{\tilde{\eta}})$ has 20 edges.}\label{Fig:filters_plot}
		\end{center}
	\end{figure}
	
	Here we present an example of our proposed filtering method to illustrate (1) how the filter produces a sparse network, (2) how different is this filtered network compared to the \emph{true} underlying network and (3) how the filtered network changes with the choice of the cost parameters. 
	
	We choose the true data-generating process to be a $p (=10)$-dimensional Gaussian distribution with mean $\bm{0}$ and covariance matrix $\Sigma$. We illustrate the method for a particular choice of $\Sigma$ (given in Appendix \ref{Appendix:true_sample_cov}). $50$ sample observations are drawn from this $p$-dimensional distribution. Therefore $p/n$ ratio is 10/50=0.2. The sample covariance matrix ($S$) is calculated from the simulated data (see Appendix \ref{Appendix:true_sample_cov}). Applying the proposed algorithm, we get the filtered network. 
	
	Fig.~\ref{Fig:true_graph} shows the true comovement network for our chosen covariance matrix $\Sigma$. Although it has a moderately sparse structure (15 undirected edges), the sample correlation matrix obtained from the simulated data is not a sparse matrix. The network constructed from the sample covariance matrix $S$, is shown in Fig.~\ref{Fig:sample_graph} (45 undirected edges indicating a fully connected network). 
	
	We plot two filtered networks corresponding to two choices of the threshold parameters.
	If we ignore the cost due to deletion of edges- i.e. if we only aim to reduce the distance between the spectral structure of the \emph{threshold}-network and the network induced by Ledoit-Wolf estimator- then we get the maximally filtered network shown in Fig.~\ref{Fig:no_cost_network}. We can see that this is not a connected network but it is able to preserve the stronger edges and the corresponding subnetworks of the true network. On the other hand, Fig.~\ref{Fig:high_cost_network} represents the filtered network obtained by introducing a positive cost, which preserves the stronger edges along with the property of connectedness.

	\begin{figure}
		\begin{center}
			\includegraphics[scale=0.6]{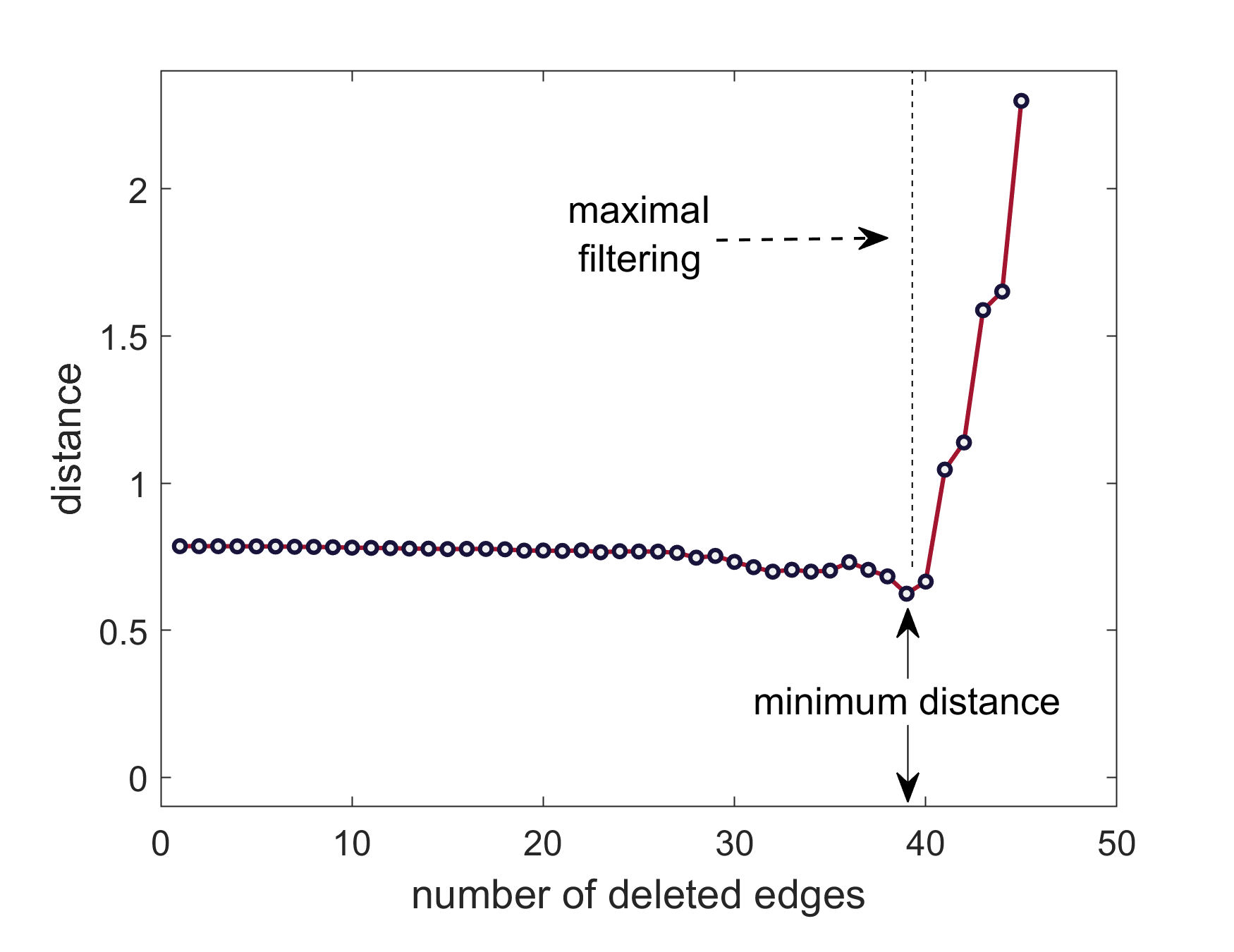}
			\caption{Illustration of the distance function against number of deleted edges corresponding to a chosen threshold- denoted by $y(\eta)$. The distance is minimized at $y = 39$ which means that the maximal filtering will remove 39 edges from $\Gamma(S)$ (the corresponding network is plotted in Fig. \ref{Fig:no_cost_network}).}
		\end{center}
		\label{fig:cost_illustration_simul}
	\end{figure}

	\subsection{Filtering with known data generating process: Information loss and spuriousness}
	\label{sec:information_loss_and}
	
	If we know the true data generating process, then tuned filtering on a sample covariance matrix gives us two informational statistics related to information loss due to edge deletion and spuriousness of edges generated by finite sampling fluctuations.
	Fundamentally, these two statistics are related to finding {\it false negatives} (deleting edges that are actually informative) and {\it false positives} (retaining edges that appear in the sample covariance matrix due to sampling fluctuation, but are not there in the true covariance matrix).
	
	The first measure we define allows us to characterize {\it true positives}. We construct the measure by the proportion of \emph{true} edges which are retained in filtered network:  
	\begin{equation}
	\text{P}_t(\eta) = \frac{\#\{E_{\Gamma(\Sigma)}\cap E_{\Gamma(S_{\eta})}\}}{\#\{E_{\Gamma({\Sigma})}\}}
	\label{eqn:p-1_eta}
	\end{equation}    
	where $\#$ denotes the cardinality of a set, $E_{\Gamma(\Sigma)}$ and $E_{\Gamma(S_{\eta})}$ are the edge sets of the true and filtered networks. The numerator and denominator of the above equation only count the number of edges and do not take into account the relative importance of the edges. The next measure replace the \emph{total number} of edges by \emph{total weight} of the edges where weight is captured by absolute value of correlation: 
	\begin{equation}
	\text{P}'_t(\eta) = \frac{\underset{(i,j)\in\{E_{\Gamma(\Sigma)}\cap E_{\Gamma(S_{\eta})}\}}{\sum} w_{i,j}}{\underset{(i,j)\in\{E_{\Gamma(\Sigma)}\}}{\sum} w_{i,j}}.
	\end{equation}  
	Clearly, both $\text{P}_t(\eta)$ and $\text{P}_t'(\eta)$ are bounded above by 1. As a consequence of being a complete graph, the sample covariance matrix has both the quantities equal to 1. The main challenge is to obtain a sparse graph with significantly high $\text{P}_t(\eta)$ and $\text{P}_t'(\eta)$. 
	
	However, high rate of edge retention might lead to retaining spurious edges.
	Therefore, it is important also to note how many edges the filtered network contains which are not part of the true network. The following proportion measures the same:   
	\begin{equation}
	\text{P}_f(\eta) = \frac{\#\{E_{\Gamma(S_{\eta})}\cap E^c_{\Gamma(\Sigma)}\}}{\#\{E_{\Gamma(S_{\eta})}\}}
	\end{equation}
	where $E^c_{\Gamma(\Sigma)}$ denotes the set of edges that do not exist in the true network (but might arise due to sampling).
	
	We report these three statistics in Table~\ref{Table:edge_prop} for different threshold parameters $\eta$ for the data generating process discussed in Sec. \ref{sec:illustrative_example}. 
	We see that when the threshold increases, $\text{P}_t(\eta)$ and $\text{P}'_t(\eta)$ decrease on average. This is intuitive because a higher threshold leads to higher number of edges being deleted and therefore, the chances of deleting true edges also go up. On the contrary, a higher threshold simultaneously 
	makes it more likely that spurious edges will be deleted. Therefore,
	the chances of having a {\it false positive} goes down. This is consistent with the column for $\text{P}_f(\eta)$ which shows that with higher threshold, the value of $\text{P}_f(\eta)$ decreases steadily.

	\begin{table}
		\begin{center}
			\begin{tabular}{ccccc}
				\toprule
				$\eta$ & $\text{P}_t(\eta)$ & $\text{P}'_t(\eta)$ & $\text{P}_f(\eta)$ \tabularnewline
				\midrule
				\midrule
				0.170 & 0.742 ($\pm$ 0.09) & 0.917 ($\pm$ 0.04) & 0.445 ($\pm$ 0.07) \tabularnewline
				0.230 & 0.698 ($\pm$ 0.09) & 0.899 ($\pm$ 0.05) & 0.351 ($\pm$ 0.09) \tabularnewline
				0.288 & 0.568 ($\pm$ 0.07) & 0.827 ($\pm$ 0.05) & 0.157 ($\pm$ 0.10) \tabularnewline
				0.499 & 0.454 ($\pm$ 0.14) & 0.752 ($\pm$ 0.08) & 0.072 ($\pm$ 0.14) \tabularnewline
				\bottomrule
			\end{tabular}\caption{Measures $\text{P}_t(\eta)$, $\text{P}'_t(\eta)$, $\text{P}_f(\eta)$ are shown for different threshold parameters $\eta$ for the data generating process discussed in Sec. \ref{sec:illustrative_example}. 
				We have computed the measures based on 100 draw of sample covariance matrices of length $n=50$ and $p=10$. The average estimates over these 100 simulations appear in the table along with the corresponding standard deviation within the adjacent brackets.
			}
			\label{Table:edge_prop}
		\end{center}
	\end{table}
	
	\section{Extensions and Robustness}
	\label{sec:choices_n_alternatives}
	
	In the following, we discuss extensions and robustness of the proposed filtering algorithm.

	\subsection{Spectral similarity in terms of subset of eigenmodes} 
	
	In the algorithm presented in section~\ref{sec:proposed_algorithm}, we have
	optimized on the threshold $\eta$ to minimize the
	distance between the spectrum of the Ledoit-Wolf estimator and the threshold estimator. However, one may not be interested in the full spectrum of covariance matrix. This is pertinent in the context of financial networks which is known to possess an eigenvalue distribution with 
	wide heterogeneity. An array of statistical analysis (see e.g. \cite{sinha2010econophysics}) shows that the highest eigenvalue captures the fluctuations due to the {\it market mode}, whereas sectoral fluctuations are associated with the deviating eigenvalues (except the largest one) from the bulk of the spectrum. The bulk of the spectrum on the other hand represents idiosyncratic fluctuations, which is modelled well by a Mar{\v{c}}enko-Pastur distribution. Therefore in this context, only the deviating eigenvalues are informative. So one can argue for considering only these few eigenvalues and choose the threshold that minimizes the distance between two vectors of deviating eigenvalues. 
	
	This represents evaluating the distance on a smaller set of eigenmodes as opposed to all eigenmodes, and the distance between eigenmodes of two matrices $A$ and $B$ to (Eqn. \ref{eq:dist_defn}) to be modified as follows:
	\begin{equation}
	d(\lambda(A),\lambda(B))=\big(\sum_{i=p_l}^{p_h}\big|\lambda_{i}^A-\lambda_{i}^B\big|^2\big)^{1/2}  \label{eq:dist_defn_subset}  
	\end{equation}
	where $1\le p_l\le p_h\le p$ and the choice of $p_l$ and $p_h$ can be chosen according to the specific system under analysis. 
	Specifically, the upper bound of the Mar{\v{c}}enko-Pastur distribution can provide such a natural cut-off for the choice of $p_h$ to capture the deviating eigenmodes and $p_l$ can be unity. 
	
	\subsection{Non-linear shrinkage estimator} 
	
	All the rotation-equivariant estimators we have discussed so far are linear. They are linear combination of the sample covariance (or correlation) matrix and a suitable shrinkage target. This means that regardless of their ranks, all the sample eigenvalues are shrunk by same intensity. However our objective is to minimize Eqn.~\ref{eq:regularization_optim} and there is no guarantee that a linear shrinkage estimator would be our best choice. \cite{ledoit2011eigenvectors, ledoit2012nonlinear} show that linear shrinkage is a first order approximation of a nonlinear problem whose utility depends very much on the situation- particularly on the limit of $p/n$. If this ratio is high then linear shrinkage will be a substantial improvement but not otherwise. Attempts have been made to find nonlinear solution to the problem which essentially results in individualized shrinkage intensity to every sample eigenvalue. First we describe the role of random matrix theory and why it is instrumental in finding the solution. 
	
	Results from random matrix theory illustrate that for high dimensional set up the eigenvalues of sample covariance matrix do not converge to its population counterparts. However, random matrix theory attempts to establish a link between the two. First attempts in nonlinear shrinkage estimators harnessed the established relation between the limiting spectral distribution of the sample eigenvalues and that of the population eigenvalues. Once the spectral distribution of the population eigenvalues are obtained, it can be numerically inverted to calculate the population eigenvalues \cite{ledoit2011eigenvectors, el2008spectrum}. Below, we describe one such solution. 
	Let us introduce the following quantities: 
	\begin{enumerate}
		\item $p/n \rightarrow c~(>0)$.
		\item If $G$ be the cumulative distribution function of eigenvalues $\lambda$, then the Stieltjes transform of the $G$ is defined as below: 
		\[
		m_G(z)=\int_{-\infty}^{\infty}\frac{1}{\lambda-z}dG(\lambda)\quad\forall z\in \mathbb{C}^+.
		\] 
		Stieltjes transform is an important tool in random matrix theory because of its one-one relationship with the distribution function (empirical spectral distribution in our context). Therefore, to determine the limiting spectral distribution one
		only needs to show the convergence of corresponding Stieltjes transform.
		\item The limiting empirical spectral distribution of the sample covariance matrix is denoted as $F$. 
		\item The Stieltjes transform of the Mar{\v{c}}enko-Pastur law \cite{bai2010spectral} is denoted by $m_F(z)$. 
		\item Define 
		\[
		\tilde{m}_F(\lambda)=\underset{z\in \mathbb{C}^+\rightarrow\lambda}{\text{lim}}m_F(z)\quad \forall \lambda \in \mathbb{R}-\{0\}
		\]
		\item Define $m_{LF}(z)=1+zm_F(z)\quad\forall z\in \mathbb{C}^+$.
	\end{enumerate}   
	Under some general assumptions, \cite{ledoit2011eigenvectors} proposed nonlinear shrinkage intensities and the derived form of $\psi$ (see Eqn. \ref{eq:stein}) is the following:
	\begin{equation}
	\psi_i = \frac{\lambda_i}{|1-c-c\lambda_i \tilde{m}_F(\lambda_i)|^2}.
	\end{equation}   
	Note that $\psi_i$ is dependent on $\lambda_i$ (unlike the linear shrinkage estimator). For more detailed discussion, see \cite{el2008spectrum, ledoit2011eigenvectors, ledoit2012nonlinear}.  
	
	There are also some other methods of nonlinear shrinkage estimation. By exploiting the connection between nonlinear shrinkage and nonparametric estimation of Hilbert transform of the sample spectral density, an analytical formula for nonlinear shrinkage has been proposed recently \cite{ledoit2020analytical}. \cite{abadir2014design} proposed a method called Nonparametric Eigenvalue-Regularized COvariance matrix estimator (NERCOME; \cite{lam2016nonparametric}) that splits the sample into two parts. One part is used to estimate the eigenvectors of the covariance matrix and the other part of sample to estimate the eigenvalues associated with these eigenvectors. Averaging over a sufficiently large number of sample split results in reasonably good estimation. All these methods can be used as alternative to the linear shrinkage estimator considered in the algorithm proposed in Sec. \ref{sec:proposed_algorithm}.       
	
	However, the difference in filtering via linear and nonlinear shrinkage estimators is often sample-dependent and exhibits large fluctuations. We attempted to filter financial networks via nonlinear shrinkage estimator (details given in Sec. \ref{sec:real_data}). Through empirical analysis, we saw that when the filtering is moderate via linear estimator, then the nonlinear estimator does not produce radically different filtering. However, we have observed extreme cases where linear shrinkage estimator leads to complete filtering of all edges whereas nonlinear shrinkage estimator led to very minor filtering. Therefore, while nonlinear shrinkage estimator has more flexibility \cite{ledoit2012nonlinear}, the corresponding impact on the strength of filtering is sample-dependent.

	\subsection{Alternative choices of the cost function} 
	
	A convex cost function captures the idea that higher number of edges being deleted would entail a higher per unit cost.\footnote{A concave cost function would lead to maximal filtering, since more filtering leads to lower per unit cost.} The underlying idea is that the first few edges being deleted would have the lowest weights. However, as we increase the threshold, the edges being filtered out would have larger and larger weights. This observation leads to the assumed convexity of the cost curve.
	The proposed functional form $C(y)=\theta_1y^{\theta_2}$ is useful for its simplicity and ease of manipulation. In principle, many other cost functions can be considered as long as they are convex in nature. A more general set of choices is presented below.

	Suppose the $(i,j)$th element of $S$ is denoted by $s_{i,j}$. We define the total edge weight as $W = \sum_{i\neq j}f(s_{i,j})$, for a suitable nonnegative function $f$. The total weight removed by filtering can be denote by $W_{\eta^*}$, which is defined as follows: 
	\begin{equation}
	W_{\eta} = \underset{(i,j)\in \bar{E} }{\sum}f(s_{i,j}),~ \text{where}~\bar{E} = E_{\Gamma(S)}-E_{\Gamma(S_{\eta})}.
	\end{equation}
	In principle, any convex function of $\frac{W_{\eta}}{W}$ is a valid choice for $C(\eta)$. As an example, if we choose $f(s_{i,j}) = I(s_{i,j} \neq 0)/2$ then $W_{\eta^*}$ becomes $y^*$ as defined in Eqn.~\ref{eq:maximal_y}.\footnote{The cost function defined in Eqn.~\ref{eq:optimal_filter_y_edge} is: 
		\begin{equation}
		C(\eta) \equiv C(y(\eta)) \equiv \theta_1 W_{\eta}^{\theta_2}.
		\end{equation} 
	} This choice of $f(.)$, although simple and easily understandable, does not directly incorporate the weight of each deleted edge. Therefore an exogenous cost function is needed to be defined and imposed (see Eqn.~\ref{eq:cost function}). An endogenous choice is the following: 
	\begin{equation}
	f(s_{i,j}) = |s_{i,j}|/2 ~ \text{and} ~ C(\eta) \propto \frac{ W_{\eta}}{W}.
	\end{equation}
	Given the above discussion, we see that this function is convex.
	
	\subsection{Alternative choices of the distance function} 
	
	In the description of the algorithm, we have utilized Euclidean distance between the spectra (Eqn. \ref{eq:dist_defn}). This is useful in terms of implementation as well as simplicity.
	In principle, one can look for alternative notions of distances as well to measure similarity or dissimilarity between two spectra. We consider them below and discuss the relative merits and demerits.

	Suppose $F(.)$ and $G(.)$ are two spectral distribution functions. For finite sample let us denote the vectors of ordered eigenvalues corresponding to two $p\times p$ matrices as $\lambda_a=(\lambda_{a,1},\lambda_{a,2},..,\lambda_{a,p})$ and $\lambda_b=(\lambda_{b,1},\lambda_{b,2},..,\lambda_{b,p})$. In this case, $F$ and $G$ are the discrete uniform distribution on $\lambda_a$ and $\lambda_b$. Three well known measures for distance are as follows:
	\begin{enumerate}
		\item \emph{Minkowski distance (for general $\kappa$)}: $d(F,G)=d(\lambda_a,\lambda_b)=\big(\sum_{i=1}^p \big|\lambda_{a,i}-\lambda_{b,i}\big|^{\kappa}\big)^{\nicefrac{1}{\kappa}}$, where $\kappa\geq1$.
		\item \emph{$L^1$ distance:} $d(F,G)=d(\lambda_a,\lambda_b)=\sum_{i=1}^p \big|\lambda_{a,i}-\lambda_{b,i}\big| $.
		\item \emph{$L^\infty$ distance:} $d(F,G)=d(\lambda_a,\lambda_b)=\text{max}_i\big|\lambda_{a,i}-\lambda_{b,i} \big|$.
	\end{enumerate}
	Note that, Minkowski distance for $\kappa=2$ is Euclidean distance which we considered in the algorithm. The other two metrics ($L^1$ and $L^{\infty}$) are special cases of the Minkowski distance. While we can potentially evaluate the filter with $L^1$ or $L^{\infty}$, given the lack of smoothness in the derivatives, we consider the $L^2$ (i.e. Minkowski distance with $\kappa$ = 2) to be the most appropriate metric for ease of computation and exposition.

	%
	%
	%
	%

	
	\section{Real-life Data Analysis}
	\label{sec:real_data} 
	
	To demonstrate application of our proposed filter, we apply it on a real-life financial data set. Although the filter would be more useful for \emph{big} data sets with high number of variables, we use a data set with a moderate number of variables for the sake of visualization of the resulting network. 
	Applicability of our method depends on three factors: 1) The data possess a  high-dimensional covariance matrix. 2) There is a network representation based on the covariance matrix. 3) There is a \emph{core} underlying set of connections or \emph{skeleton} of the network that is captured by large enough covariances.
	
	\subsection{Application to financial networks}

\begin{figure}
		\begin{center}
			\begin{minipage}{0.30\textwidth}
				\includegraphics[trim=15 80 16 16,clip,width=\textwidth]{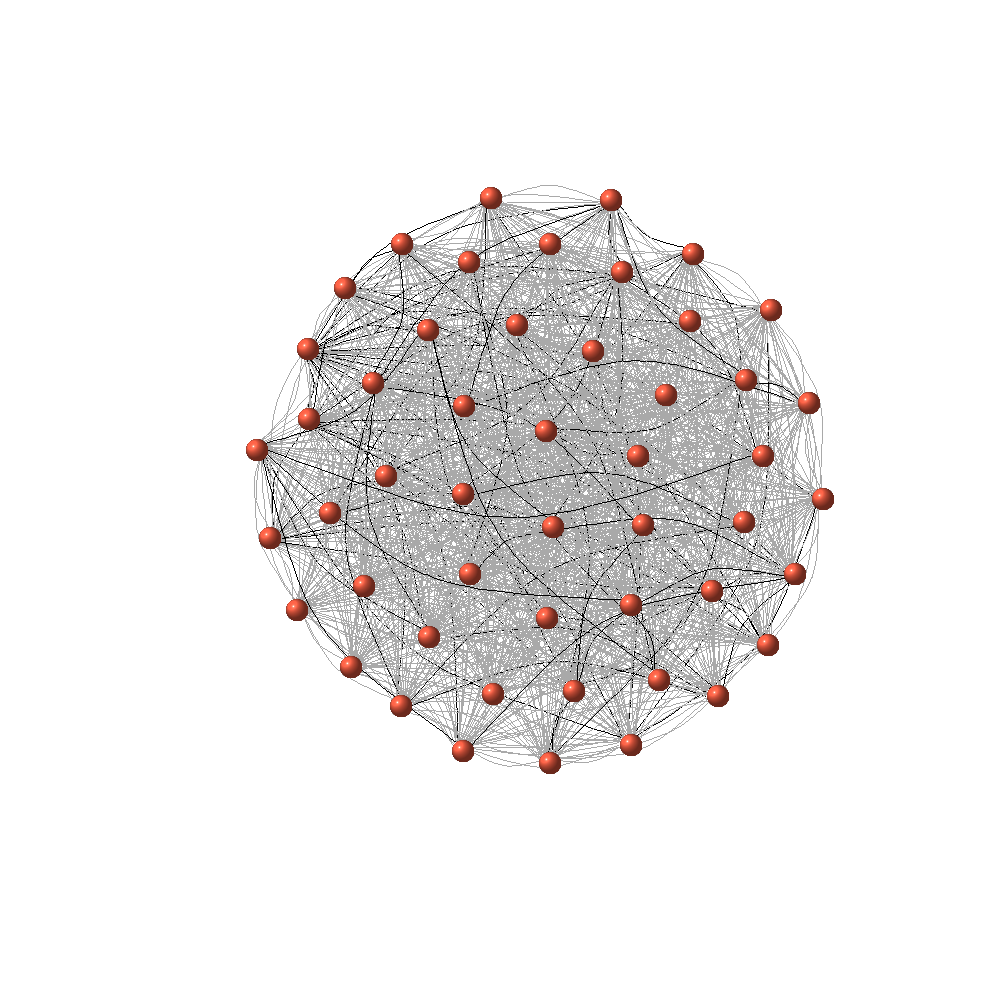}
				\vspace{.1 cm}
				\subcaption{Financial network}\label{Fig:real_data_sample_graph}
			\end{minipage}
			\begin{minipage}{0.30\textwidth}
				\includegraphics[trim=15 80 16 16,clip,width=\textwidth]{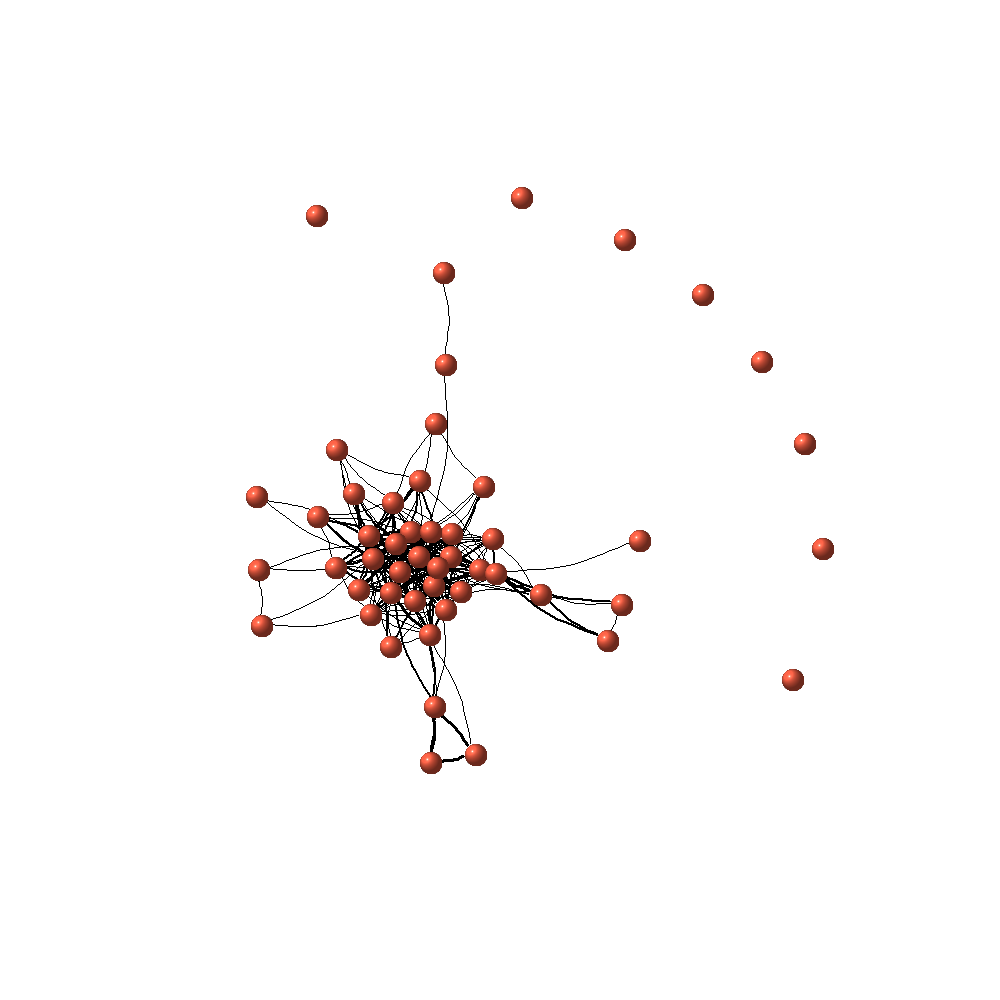}
				\vspace{.1 cm}
				\subcaption{
					Maximally filtered network
				}\label{Fig:real_data_filtered}
			\end{minipage}
			\begin{minipage}{0.27\textwidth}
				\includegraphics[trim=15 0 16 16,clip,width=\textwidth]{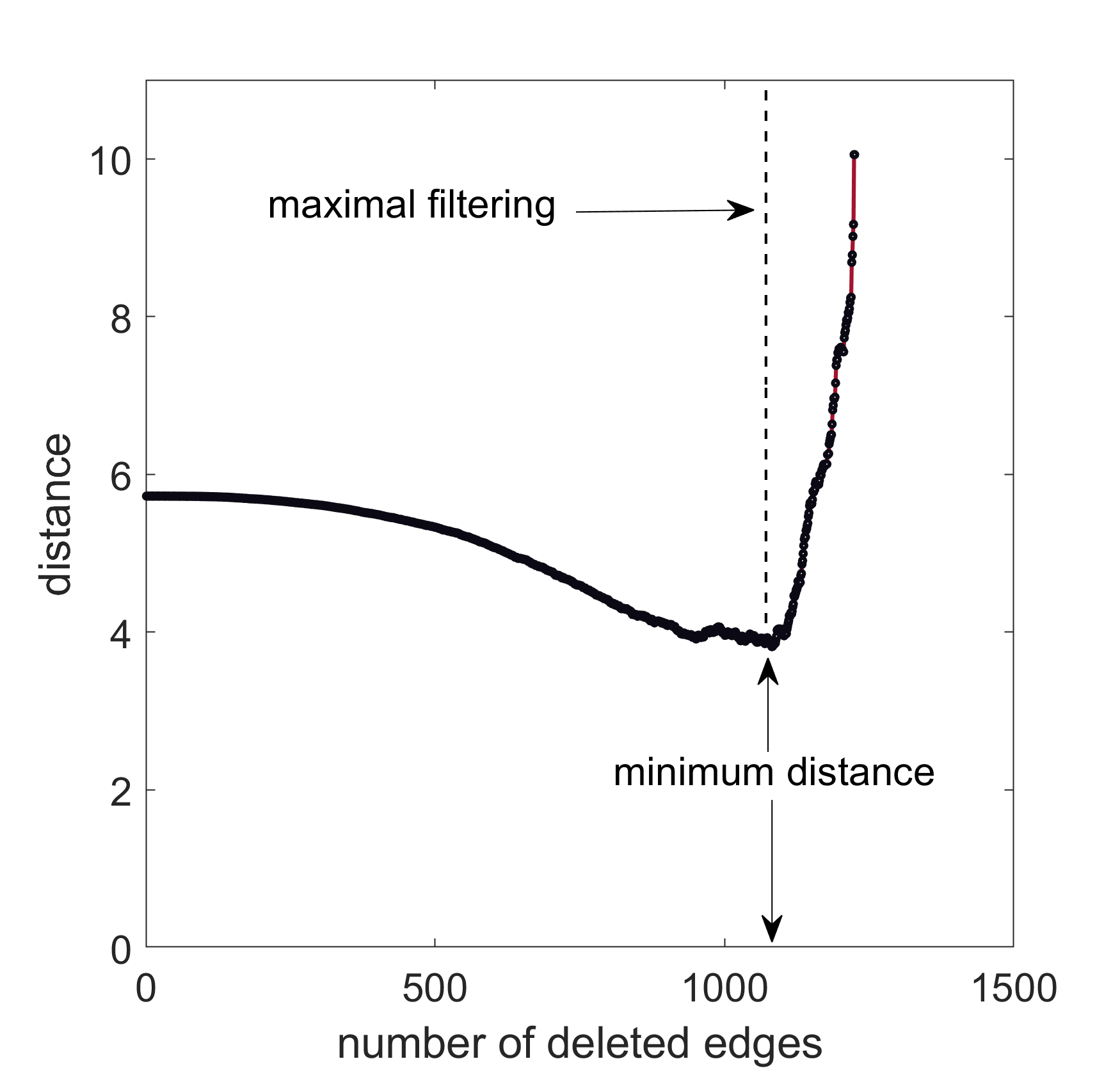}
				\subcaption{Spectral distance}\label{Fig:real_data_spectral_dist}
			\end{minipage}
			\vspace{.3 cm}
			\caption{Illustration for Financial network. Panel (a) shows the network ($\Gamma(S)$) obtained from sample correlation matrix. It is a complete graph with dense connectivity. Panel (b) shows the maximally filtered network with substantially fewer edges. The network splits into a giant connected component and unconnected peripheral nodes. Panel (c) shows the spectral distance function against number of deleted edges. The lowest distance corresponds to the maximal filtered network shown in panel (b).}
		\end{center}
	\end{figure}

	We perform our method on historical NASDAQ data for 50 stocks with prices recorded over 70 consecutive days, from 2nd January, 2015 to 14th April, 2015. As is customary in the analysis of return comovement networks \cite{sinha2010econophysics}, we first construct the log return series for each of the stocks. If a stock's price at time $t$ is $P_t$, then the log return at time $t$ is defined as the following:
	\begin{equation}
	r_t=\text{log} P_t - \text{log} P_{t-1}.
	\end{equation}
	From the generated return series, the sample correlation matrix $S$ is calculated and the corresponding network $\Gamma(S)$ is generated (Fig.~\ref{Fig:real_data_sample_graph}). As it is a complete graph we can see all possible edges are present (we have excluded the self-loops) as all pairs of stocks would exhibit non-zero covariance. 
	
	Fig.~\ref{Fig:real_data_filtered} exhibits the maximally filtered network, which shows a drastic reduction in the number of edges(from 1225 to 256). 
	We also observe that this filtered graph is not connected. In particular, the maximally filtered network produces 8 isolated vertices while the remaining 42 stocks create a giant component.\footnote{In some sets of stocks, we have observed that the spectral distance between the true matrix and the Ledoit-Wolf analog is so large that the maximal filtering leads to fully diagonal matrix, i.e. all nodes become separate. In such cases, the spectral similarity is not an useful criterion for filtering.}

	%
	
	

	Finally, in Fig. \ref{Fig:real_data_spectral_dist} we plot the spectral distance between the Ledoit-Wolf estimator and the filtered matrices in an increasing number of edges being deleted (associated with increasing threshold). The global minimum for the distance function is reached  
	at 1082 (=1225-143) number of edges being deleted. Therefore, the maximally filtered network would consist of 143 edges as shown in Fig. \ref{Fig:real_data_filtered}. Clearly, such strong filtering is associated with high thresholding and consequent loss of a large number of edges. For a less strong filtering, one can impose a positive cost of edge deletion (maximally filtering requires zero cost of edge deletion) following Eqn. \ref{eq:cost function}. based on the choice of parameters, one can interpolate between zero filtering to maximal filtering.

	To complement the above analysis, we carry out the filtering on a large covariance matrix arising out of the largest $p$ =  300 stocks in NASDAQ in 2015 calendar year by varying the number of observations $n$ from 50, 200, 300 and 450. The values are chosen such that the $p/n$ ratio varies from a number smaller than one to larger than one. The resulting maximal thresholds are shown in Fig. \ref{fig:finacial_example_large_15_18} in the Appendix \ref{Appendix:fin_ex_2015_18}. As can be seen, for large ratio of $p$ to $n$ (indicating very small number of observations for each stock) the filtering threshold is very high and a high fraction of edges get filtered. In the other extreme, when $p$ to $n$ ratio is small (indicating a large number of observations for each stock), then the filtering threshold is very low and therefore, very few edges are filtered. For the sake of completeness, we should mention that the filtering threshold is influenced by sampling fluctuations and therefore, such a monotonic relationship may not be found in all applications. However,
	in our numerical experiments we found that on an average a larger number of observations for each entity (stocks in this case) leads to smaller filtering threshold.

	\section{Summary and Conclusion}
	\label{sec:summary}
	
	Many large scale systems are best described as networks \cite{barabasi2004network,barfuss2016parsimonious,battiston2016complexity,cho2012network,huberman1999growth,newman2003structure,marti2017review,sinha2010econophysics}. A standard approach of network construction is to create covariance-based measures of interlinkages \cite{sinha2010econophysics}.
	However, construction of the comovement network from an observed data set is a challenging problem because the resulting network is a complete graph and therefore resists any naive attempt to uncover the underlying network topology due to existence of spurious linkages. Statistically such networks suffer from false positives, i.e. false discovery of linkages. Therefore,
	a robust methodology is needed to identify and prune such non-informative linkages and isolate the key subnetwork embedded in the complete network. In this paper, we develop a filtering technique that attempts to resolve this problem utilizing spectral structure of the network.
	
	The existing filtering techniques have mainly two features. First, they are primarily based on some graph-theoretic constraints and not on explicit statistical motivation 
	(e.g. minimum spanning tree or more general, hierarchical structures). Second, many of the filtering techniques are not tunable and often they lead to a drastic reduction in the number of edges (e.g. minimum spanning tree), which also makes the resulting network very unstable and sample-dependent. In this paper, we propose an new filter based on the properties of high-dimensional covariance estimators, utilizing the concept of {\it sparsistence} along with retaining flexibility for tuning the degree of filtering.
	We note that one can consider algorithms based on hypothesis testing of individual edge weights and prune statistically insignificant edges. However, this kind algorithms still suffer from the problem of false positives (i.e. false discovery of edges) as they do not account for joint hypothesis testing.

	We approach the problem in a new way by considering the spectral structure of the covariance network and sparsistent analogues of that, based on Ledoit-Wolf estimators which features predominantly in high dimensional covariance matrix estimation.
	Depending on the statistical properties of the Ledoit-Wolf estimator, we prescribe an  endogenously thresholded covariance matrix estimator such that its spectrum is closest to that of the Ledoit-Wolf matrix. 
	We complement the theoretical structure with numerical simulations along with 
	applications to real world financial data.
	
	Our work is situated in the intersection of the literature on network filtering, covariance matrix estimation and large dimensional data. The proposed algorithm can be applied to any large dimensional data. We have demonstrated the usefulness of the filtering algorithm by applying it to financial stock return data. Further applications to various domains spanning biological, physical and technological comovement networks would lead to a more complete understanding of the corresponding topological structures and key linkages that contribute to the dynamics of the system.

	\clearpage
	\bibliographystyle{plain}

	\clearpage
	\newpage
	\appendix
	\section{Appendix}

	\subsection{True and sample covariance matrix for the illustrative example}\label{Appendix:true_sample_cov}
	The true correlation matrix for the simulated illustrative example (Sec.~\ref{sec:illustrative_example}) is the following: 
	
	\begin{center}
		\[
		\begin{bmatrix}
		1.00  & 0.80  & 0.00  & 0.00  & 0.00  & 0.30  & 0.00  & 0.00  & 0.00  & 0.00\\
		0.80 & 1.00 & 0.00 & 0.00 & 0.00 & 0.09 & 0.00 & 0.00 & 0.00 & 0.00\\
		0.00 & 0.00 & 1.00 & 0.30 & 0.09 & 0.09 & 0.00 & 0.09 & 0.00 & 0.00\\
		0.00  & 0.00  & 0.03  & 1.00  & 0.00  & 0.00  & 0.30  & 0.00  & 0.00  & 0.00\\
		0.00 & 0.00 & 0.09 & 0.00 & 1.00 & 0.80 & 0.00 & 0.00 & 0.00 & 0.00\\
		0.30 & 0.09 & 0.09 & 0.00 & 0.80 & 1.00 & 0.00 & 0.00 & 0.00 & 0.00\\
		0.00  & 0.00  & 0.00  & 0.30  & 0.00  & 0.00  & 1.00  & 0.30  & 0.80  & 0.80\\
		0.00 & 0.00 & 0.09 & 0.00 & 0.00 & 0.00 & 0.30 & 1.00 & 0.09 & 0.30
		\end{bmatrix}
		\]
	\end{center} 
	After generating the sample, the sample covariance matrix is: 
	\begin{center}
		\[
		\begin{bmatrix}
		14.36 & -2.12 & 2.36  & -1.44 & 0.38  & -4.01 & 0.64  & 2.65  & -3.38 & -2.13\\
		-2.12 & 7.02  & -0.63 & -0.01 & -0.80 & 1.20  & 0.12  & -3.29 &  1.87 & -5.86\\
		2.36  & -0.63 & 11.12 & 5.55  & -0.06 & -1.31 & 5.04  & 4.63  & -3.83 & 2.14\\
		-1.44 & -0.01 & 5.55  & 9.73  & -3.03 & -1.69 & 7.45  & 1.15  & -4.97 & 1.85\\
		0.38  & -0.80 & -0.06 & -3.03 & 6.76  & 3.57  & -2.23 & -3.82 & -2.84 & -0.96\\
		-4.01 & 1.20  & -1.31 & -1.69 & 3.57  & 5.99  & -4.84 & -2.91 & -2.08 & -1.12\\
		0.64  & 0.12  & 5.04  & 7.45  &-2.23  & -4.84 & 13.54 & -1.80 & -3.02 & 0.33\\
		2.65  & -3.29 & 4.63  & 1.15  & -3.82 & -2.91 & -1.80 & 9.56  & 1.38  & 4.99
		\end{bmatrix}
		\]
	\end{center} 
	
	We see that many entries of the true correlation matrix is 0 and therefore, the corresponding covariance matrix would be a sparse matrix. However, it is noteworthy that the sample covariance matrix does not contain any 0 due to sampling fluctuations. So the resulting network representation will be a fully connected network although the underlying network is sparsely connected.

	\subsection{Application on large dimensional financial covariance matrix}
	\label{Appendix:fin_ex_2015_18}

	\begin{figure}
		\begin{center}
			\includegraphics[trim=0 0 0 0,clip,width=\textwidth]{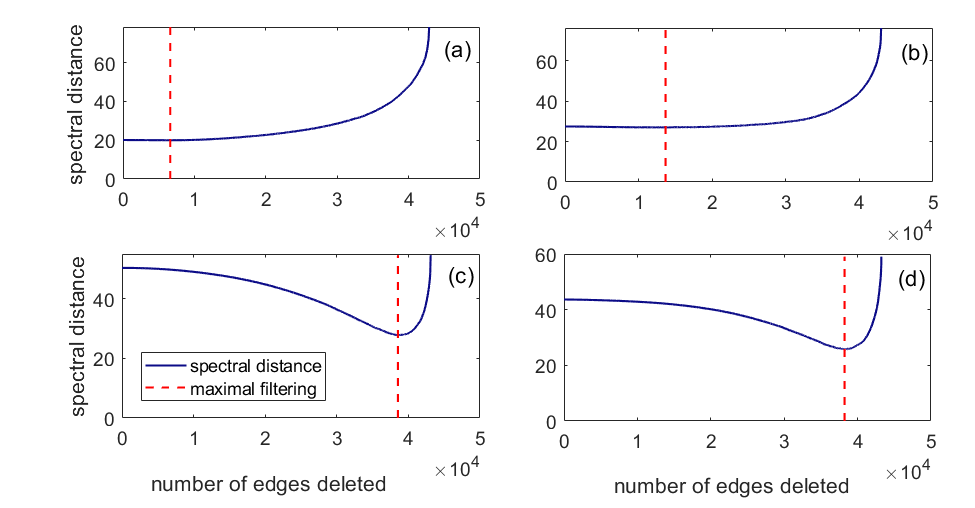}
			\vspace{.1 cm}
			\caption{Application of the sparsistent filter on large dimensional financial network. We have considered the covariance matrices of $p=300$ largest stocks from NASDAQ stock exchange (in terms of market capitalization) with varying number of days. Each data set begins on 2nd January, 2015 and continues for $n$ days where $n$ varies from 50, 200, 300 and 450. Therefore, in Panel (a) $p/n$ = 2/3, Panel (b) $p/n$ = 1, Panel (c) $p/n$ = 3/2 and finally, Panel (d) $p/n$ = 6. A larger $p/n$ ratio leads to larger threshold for maximal filtering.}
			\label{fig:finacial_example_large_15_18}
		\end{center}
	\end{figure}

\end{document}